\pdfoutput=1

\documentclass[11pt]{article}

\usepackage{emnlp2021}

\usepackage{times}
\usepackage{latexsym}
\usepackage{enumitem}
\usepackage[T1]{fontenc}

\usepackage[utf8]{inputenc}

\usepackage{microtype}

\usepackage{subcaption}
\usepackage{booktabs}
\usepackage{tabularx}
\usepackage{arydshln}
\usepackage{multirow}
\usepackage{graphicx}
\usepackage{array, makecell}

\title{Rome was built in 1776: A Case Study on Factual Correctness in Knowledge-Grounded Response Generation}

  \author{ Sashank Santhanam\textsuperscript{{\rm 1}} \quad Behnam Hedayatnia\textsuperscript{{\rm 2}} \quad  Spandana Gella\textsuperscript{{\rm 2}} \\ \textbf{Aishwarya Padmakumar\textsuperscript{{\rm 2}} \quad Seokhwan Kim\textsuperscript{{\rm 2}} \quad  Yang Liu\textsuperscript{{\rm 2}} \quad Dilek Hakkani-Tur\textsuperscript{{\rm 2}}} \\
  \textsuperscript{\rm 1}University of North Carolina, Charlotte, 
\textsuperscript{\rm 2}Amazon Alexa AI\\
\texttt{ssantha1@uncc.edu}\\  
\texttt{\{behnam,sgella,padmakua,seokhwk,yangliud,hakkanit\}@amazon.com}\\ 
  
  \\}

\begin{document}
\maketitle
\begin{abstract}

Recently neural response generation models have leveraged large pre-trained transformer models and knowledge snippets to generate relevant and informative responses. However, this does not guarantee that generated responses are factually correct. In this paper, we examine factual correctness in knowledge-grounded neural response generation models. We present a human annotation setup to identify three different response types: responses that are factually consistent with respect to the input knowledge, responses that contain hallucinated knowledge, and non-verifiable chit-chat style responses. 
We use this setup to annotate responses generated using different state-of-the-art models, knowledge snippets, and decoding strategies. In addition, to facilitate the development of a factual consistency detector, we automatically create a new corpus called Conv-FEVER that is adapted from the Wizard of Wikipedia dataset and includes factually consistent and inconsistent responses.
We demonstrate the benefit of our Conv-FEVER dataset by showing that the models trained on this data perform reasonably well to detect factually inconsistent responses with respect to the provided knowledge through evaluation on our human annotated data. We publicly release the human annotated responses~\footnote{https://github.com/alexa/factual-consistency-analysis-of-dialogs}.

\end{abstract}

\begin{figure}[!ht]
\centering
\begin{tabular}{p{0.95\linewidth}}
\hline
{\em Dialog History:}\\
{\bf Speaker 1:} What do you think about Murray? \\
{\bf Speaker 2:}  I think Murray is a great player he just needs to stay healthy in order to compete more. Who do you like best? \\ 
\hline
{\em Knowledge}: \\
In tennis, the term Big Four refers to the quartet of men's singles players comprising Roger Federer, Rafael Nadal, Novak Djokovic, and Andy Murray.\\
\hdashline[.4pt/1pt]
{\em Non-verifiable Response}:\\ 
I like Federer and Nadal. I think Federer is the best.\\
\textcolor{blue}{\textbf{Verifiable}} = No; \textcolor{orange}{\textbf{F.C}} = N/A; \textcolor{red}{\textbf{H}} = N/A \\
\hdashline[.4pt/1pt]
{\em Factually Consistent Response:} \\
Rafael Nadal is my favorite of the Big Four. \\
\textcolor{blue}{~\textbf{Verifiable}} = Yes; \textcolor{orange}{\textbf{F.C}} = Yes; \textcolor{red}{\textbf{H}} = No \\
\hdashline[.4pt/1pt]
{\em Hallucinated Response:} \\
I like Djokovic.~\textcolor{red}{\textbf{He has played in the top ten singles players of the world.}}\\
\textcolor{blue}{~\textbf{Verifiable}} = Yes; \textcolor{orange}{\textbf{F.C}} = Yes; \textcolor{red}{\textbf{H}} = Yes \\
\hline
\end{tabular}
\caption{{\em Non-verifiable Response} does not include any information that needs to be verified and cannot be evaluated as consistent or not consistent.~{\em Factually Consistent Response} is consistent with the provided knowledge.~{\em Hallucinated Response} is not consistent with the knowledge but may still be correct. \textcolor{orange}{\textbf{F.C}} = Factual Consistency. \textcolor{red}{\textbf{H}} = Hallucination. 
Description of the annotation for these labels are in Section~\ref{sec:human_annotation}.}
\label{fig:problem_set}
\end{figure}

\section{Introduction}
\label{sec:introduction}
One of the biggest challenges in the field of conversational AI is to develop agents that can generate coherent and grammatical responses whilst also seamlessly blending in knowledge to produce informative responses \citep{roller2020open}. With the development of large scale language models, conversational AI systems have made significant progress in being able to generate coherent and grammatical text. To produce informative responses, prior neural response generation work proposed grounding on knowledge sentences that are relevant to the dialog context~\citep{ghazvininejad2018knowledge,yavuz2019deepcopy, zhou2018commonsense,roller2020recipes}. However, it is not guaranteed that these models generate responses that leverage the provided knowledge and are factually correct. Being able to generate factually correct responses is important because providing incorrect information can reduce response quality and can even be critical especially for certain domains such as politics, medicine, or finance.

Previous studies have evaluated a neural model's factual correctness in several fields. 
For example, for large pre-trained models, 
it is shown that these generative models are able to memorize knowledge, but still generate factually incorrect responses ~\citep{petroni2019language, logan2019barack, roberts2020much}.

Factual correctness has been studied in multiple generation tasks such as summarization~\citep{maynez-etal-2020-faithfulness, lux2020truth, kryscinskiFactCC2019, nan2021improving, zhou2020detecting}, document-level generation~\citep{massarelli2019decoding} and natural language generation (NLG)~\citep{duvsek2020evaluating, thomson2020gold}. While factual correctness has been studied in knowledge-grounded response generation ~\citep{li2019incremental, shuster2021retrieval, honovich2021q, rashkin2021increasing}, there has been very limited work in creating a human annotated dataset and detecting factual consistency for open-domain dialog systems.

In this work, we conduct a thorough study to evaluate how factually consistent neural response generation models are with respect to the provided knowledge sentence or text retrieved from a knowledge base.  
We include a variety of GPT2-based models with different sizes, decoding strategies and quality of knowledge sentences. For our annotation of factual correctness, we categorize responses into three types as shown in Figure~\ref{fig:problem_set}: non-verifiable responses that don't contain information that need to be verified, factually consistent responses with respect to the input knowledge,  and hallucinated responses that contain information not found in the input knowledge. 
Such annotations are different from previous work in that we are dealing with open-domain dialogs, where there are responses that don't need to be evaluated for factual correctness. We form two datasets from these annotations: a large annotated dataset from public crowd-workers that labeled generated responses across different knowledge retrieval, decoding and model setups, and a smaller annotated dataset from internal crowdworkers to improve annotation quality.

After showing that state-of-the-art neural response generation models do produce factually inconsistent responses, we propose to build a factual consistency detection model. To train such a model, we created a corpus, Conv-FEVER, that includes the human written responses in the  Wizard of Wikipedia dataset~\citep{dinan2018wizard} and automatically generated factually inconsistent responses.
We demonstrate competitive performance of our detector on human annotated data. 
Note that we make a distinction between factually~\textbf{consistent} and factually~\textbf{correct} responses.
The former accurately portrays the input knowledge (assuming the provided knowledge input is correct), and the latter is accurate with respect to the ``world knowledge". 
Our detection model focuses specifically on factual consistency. Checking if a response is correct against ``world knowledge" is an important problem that we leave for future exploration.

In summary, this is the first study with a thorough analysis of factual correctness for knowledge-grounded neural response generation models. In addition to the Conv-FEVER corpus, we are also releasing our annotated corpus of factual correctness on responses from multiple neural response generation models. We expect this study will increase awareness of the factual correctness problem in knowledge-grounded dialog systems and these datasets can help benchmark progress in this area.

\section{Related Work}
\label{sec:related_work}

Previous work in detecting factual consistency have leveraged natural language inference datasets and models. \citet{maynez-etal-2020-faithfulness} used a model trained on MNLI to detect consistency between abstractive summaries and their source documents. \citet{duvsek2020evaluating} also applied a model trained on MNLI to detect hallucinations and omissions for data-to-text generation. Howeve, the MNLI dataset consists of sentences from a document while our work focuses on conversations.

Different decoding mechanisms have also been explored to improve factual correctness in generative models. \citet{massarelli2019decoding} found that delayed beam-search helps increase factual correctness for document level generation. \citet{li2019incremental} proposed a two-pass approach to generate factually consistent responses for knowledge-grounded response generation. \citet{dziri2021neural} queried a knowledge graph using an entity retrieval model to improve factual correctness in response generation. \citet{shuster2021retrieval} studied how different knowledge retrieval setups affect factual correctness in generated responses. However, their work focused on comparing factual correctness in non-knowledge grounded versus knowledge-grounded response generation models. Our work focuses purely on knowledge-grounded response generation. Additionally, we take into account the quality of the provided knowledge and annotate responses using different knowledge retrieval setups. 

There has also been previous work on generating synthetic data to train a factual consistency detector. \citet{kryscinskiFactCC2019} generated factually incorrect examples by using techniques such as entity swapping and negation. This data was used to train a factual detector on generated summaries. \citet{Schuster2019} generated negative examples for sentence level verification. We look at generating factually incorrect examples as well; however, our work focuses on factual consistency for the conversation domain.

Previous work also includes studies on human evaluation setups for  factual correctness. \citet{thomson2020gold} defined a methodology to annotate factual correctness for the data-to-text task. \citet{mielke2020linguistic} proposed an annotation schema for the correctness and confidence of a model's response, and calibrated the response such that its correctness aligns with its confidence. However, their work focused on a Q/A dataset and on knowledge that a model has learned during training. 

Most recently, \citet{honovich2021q} proposed to use question generation/answering to evaluate factual consistency for knowledge-grounded dialogs. They annotated two model outputs for factual correctness. Within their annotation schema, they asked evaluators to ignore responses that are uninformative or chit-chat. Our annotation schema is similar in that we are also annotating if responses need to be verified before annotating for factual consistency. However, our dataset includes annotations for a diverse set of responses, generated across different knowledge retrieval, decoding and model setups to demonstrate the performance of different systems. We will be releasing our human annotations that consists of both if a sentence is verifiable and if it is factually consistent or hallucinates knowledge. We also propose to use different factual consistency detection approaches.

\begin{table*}[t]
 \begin{tabular}{p{0.1\textwidth} p{0.5\textwidth} p{0.3\textwidth}}
  \toprule
 Type & Context & Response \\
 \toprule
  \multirow{2}{0.1\textwidth}{Opinion} & \textbf{B}: Ketchup and potato chips go well together too. & \multirow{3}{0.3\textwidth}{I like poutine and chili cheese fries too.} \\ 
  & \textbf{A}: I like the crinkle cut fries too. & \\ \midrule
  \multirow{2}{0.1\textwidth}{Suggestion}  & \textbf{A}: If everybody liked the same things the world would be such a dull place. & \multirow{2}{0.3\textwidth}{I agree, and I'm sure you'd enjoy a good cup of coffee!} \\ 
  & \textbf{B}: I'm glad our differences can make the world interesting.  & \\ \midrule
  Topic Switch  & \textbf{A}: i grew up in n.y. so i'm familiar with all that & I would love to visit the museums there. \\ \midrule
  \multirow{3}{0.1\textwidth}{Yes-No Question}  & \textbf{B}: How do you usually make pork steak, do you broil or pan fry them?  & \multirow{3}{0.3\textwidth}{Do you like to flip them at all?} \\ 
  & \textbf{A}: I always fan fry them. & \\ \midrule
  Personal information  & \textbf{A}:  Yes, in my area winter has a lot of snow and freezing temperatures.  Do you have a cold winter where you are? & Yes, I do.  I live in the Arctic region. \\ \bottomrule
 \end{tabular}
 \caption{Examples of different non-verifiable responses.
 \label{chitchat_examples}}
\end{table*}

\section{Are Neural Responses Consistent With Grounded Knowledge?}
\label{sec:factual_issues}
Knowledge-grounded dialog systems leverage knowledge snippets to have engaging conversations with users.
In this study, we first evaluate how factually consistent the responses are in such systems.  
We identify three types of responses within open-domain dialogs, as shown in Figure~\ref{fig:problem_set}.

\paragraph{Non-verifiable responses} do not leverage the knowledge in a way that it needs to be verified. In Figure~\ref{fig:problem_set}, the {\em Non-verifiable Response} extracts the entities \textit{Federer} and \textit{Nadal} from the  {\em Knowledge}, but expresses the information about the entities as an opinion. Table~\ref{chitchat_examples} shows more examples of different types of non-verifiable responses.

\paragraph{Factually consistent responses} contain detailed information and do not contradict the {\em Knowledge}. In Figure~\ref{fig:problem_set}, the response implies that \textit{Rafael Nadal} is a part of the Big Four. This requires verification to see if this information is valid based on the given knowledge.

\paragraph{Hallucinated responses} are not consistent with the information provided in the knowledge. 
In Figure~\ref{fig:problem_set}, the {\em Hallucinated Response} example leverages part of the {\em Knowledge} by including the entity ~\textit{Djokovic}, but also hallucinates information (shown in the highlighted text). This piece of information is not present in the knowledge, and is not clear if it is correct. 
Some of the hallucination may be correct with respect to the ``world knowledge". 
This is similar to the hallucination problem in summarization and NLG tasks~\citep{thomson2020gold}. 

Responses can also contain a mixture of verifiable and non-verifiable spans. For the {\em Hallucinated Response} in Figure~\ref{fig:problem_set}, the first sentence is an opinion on the entity while the second half is a factual statement on the same entity (not based on the knowledge sentence). 

\subsection{Annotation Scheme}
\label{sec:human_annotation}
We propose a two-stage human annotation setup for factual correctness in the context of open-domain agents.
The annotators are given a dialog context along with a knowledge sentence, and evaluate the system generated response. Stage 1 involves evaluating~\textbf{Appropriateness} and~\textbf{Verifiability} on a Likert scale of 1-5. For this stage we only show the dialog context and response. We define appropriateness as relevance to the dialog context and verifiability as to what degree the response needs to be verified. This stage is used to filter out incoherent and non-verifiable responses. If a response scores low on verifiability, it is categorized as a non-verifiable response.

Stage 2 of our setup involves evaluating~\textbf{Factual Consistency} and~\textbf{Hallucination}. We pose the Factual Consistency question: ~\textit{Is the response generated factually accurate with regards to the input knowledge?} with a three-point scale: factually incorrect (0), partially correct (0.5), and completely correct (1).
For Hallucination, we pose the question: \textit{Is the response generated making up more information than what is provided in the conversational context and input knowledge?}, with a binary label (Yes/No) for whether each response contains any hallucinated information. We present the annotation interface  used for this collection in the Appendix. 
The example in Figure~\ref{fig:problem_set} shows the annotation for verifiability, factual consistency, and hallucination, for each type of responses.

\begin{table*}[ht]
\centering
\tabcolsep 2.5pt
\begin{tabular}{cc c ccc c ccc c ccc c ccc }
    \toprule
    \multirow{3}{*}{Retriever} & \multirow{3}{*}{Decoding} && \multicolumn{3}{c}{GPT2-Small} && \multicolumn{3}{c}{GPT2-Medium} && \multicolumn{3}{c}{GPT2-Large} && \multicolumn{3}{c}{GPT2-XL}\\ 
    \cline{4-6} \cline{8-10} \cline{12-14} \cline{16-18}
    & 
    && F$~\uparrow$ & H$~\downarrow$ & A$~\uparrow$ 
    &&F$~\uparrow$ & H$~\downarrow$ & A$~\uparrow$ 
    &&F$~\uparrow$ & H$~\downarrow$ & A$~\uparrow$ 
    &&F$~\uparrow$ & H$~\downarrow$ & A$~\uparrow$\\ 
    \cline{1-2} \cline{4-6} \cline{8-10} \cline{12-14} \cline{16-18}
    \multirow{3}{*}{GT} & BS && 0.78 & 17.2\% & 3.9 && 0.80 & 18.4\% & 3.7 && 0.82 & 19.1\% &\textbf{4.1} && 0.84 & 8.0\% & 3.8\\
    & NS && 0.80 & 19.0\% & 3.8& & 0.84 & 11.6\% & 4.0 &&\textbf{0.88} & 8.5\% & 4.0 && \textbf{0.88} & \textbf{3.2\%} & 3.9 \\
    & DBS && 0.81 & 22.2\% & 3.8 && 0.82 & 13.8\% & 4.0 && 0.81 & 11.7\% &\textbf{4.1} && 0.83 & 8.0\% & 3.8\\
    \hdashline[.4pt/1pt]
    \multirow{3}{*}{KNN} & BS && 0.72 & 25.4\% & 3.6 && 0.74 & 21.7\% & 3.5 && 0.72 & 28.0\% & 3.6 && 0.77 & 22.7\% & 3.7\\
    & NS && 0.70 & 32.5\% & 3.4 && \textbf{0.83} &~\textbf{16.0\%} & 3.6 && 0.74 & 17.5\% &\textbf{3.8} && 0.75 & 28.0\% & 3.7\\
    & DBS && 0.71 & 21.4\% & 3.5 && 0.69 & 20.0\% & 3.7 && 0.70 & 28.3\% & 3.6 && 0.70 & 22.0\% &\textbf{3.8}\\
    \hdashline[.4pt/1pt]
    \multirow{3}{*}{DPR} & BS && 0.65 & 30.5\% & 3.1 && 0.64 & 26.8\% &\textbf{3.5} && 0.69 & 30.7\% & 3.4 && 0.62 & 35.1\% & 3.4 \\
    & NS && 0.61 & 43.2\% & 3.1 && 0.65 & 30.2\% & 3.4 && 0.7 & 24.4\% &\textbf{3.5} && 0.67 & 27.0\% & 3.4 \\
    & DBS && 0.59 & 39.0\% & 3.2 && 0.62 & 35.6\% &\textbf{3.5} && 0.7 & 24.4\% & 3.4 && \textbf{0.71} & \textbf{18.2\%} & 3.3 \\ 
    \bottomrule 
\end{tabular}
\caption{\label{human_all_knowledge}
Human annotation results on data {\it Factual Dataset-crowd}. F = Factual Consistency score (0-1), H = Hallucination score (\% of responses labeled ``Yes''). A = Appropriateness (1-5)}
\end{table*}

\subsection{Factual Correctness Study on Neural Response Generators}
\label{sec:data_annotation}
To quantitatively present the issues of factual correctness in open-domain dialog systems, we leverage our human annotation setup to label a set of outputs from different neural response generators. We use public crowd-workers in order to annotate a large amount of data under different configurations, namely model size, knowledge retrieval, and decoding setups. We vary knowledge retrieval setups as factual correctness in a response can vary depending on how accurate a knowledge retrieval model is~\citep{schuster2021get}. We denote this dataset as~\textit{Factual dataset-crowd}.

We leverage Wizard of Wikipedia (WoW)~\citep{dinan2018wizard}, a knowledge grounded dialog dataset generated through MTurkers who played the role of wizard and apprentice. The wizard has access to Wikipedia passages and the apprentice is given the role of learning more about a topic by engaging in a dialog with the wizard. At every turn the wizard selects a knowledge sentence from the Wikipedia passages to generate a knowledge-grounded turn.

For the response generation models, we leverage GPT2~\cite{radford2019language} and follow the end-to-end-approach of~\citep{dinan2018wizard, gopalakrishnan2019topical} and take in as input both the dialog context and a knowledge sentence and minimize cross-entropy loss on the ground-truth response. 
We fine-tune our models in a TransferTransfo fashion~\citep{wolf2019transfertransfo} on the WoW dataset. The dialog context and the knowledge sentence are represented with the pretrained embeddings from the GPT2 model whose vocabulary is BPE tokenized. Hyperparameters are provided in the Appendix.

To evaluate factual consistency of different response generators, we vary the generator from three aspects: GPT2 model size, provided knowledge, and decoding strategy.  

    \paragraph{Model size:} We use the four variants of GPT2~\citep{radford2019language} based language models: small, medium, large, XL.
    \paragraph{Decoding strategies:} For decoding we use Nucleus sampling (NS)~\citep{holtzman2019curious} with p=0.9; Beam-Search (BS) with a beam size of 5, and Delayed Beam-Search (DBS)~\citep{massarelli2019decoding}, which uses top-k sampling for n delay steps followed by beam search. We use k=10, 5 delay steps and a beam size of 5. 
    \paragraph{Knowledge:}  we use three types of configurations for the knowledge sentence:

        \paragraph{(a) Ground Truth knowledge (GT):} This is the knowledge used by the ``Wizards'' to generate their response. It is explicitly available in the WoW test set and it is also referred to as the checked sentence in the WoW data.
        
        \paragraph{(b) K-Nearest Neighbours (K-NN)}: In this configuration, we use a $K$ nearest neighbor search across all the 5 million Wikipedia articles used in the WoW dataset. In order to construct the knowledge base \textbf{(E)}, we encode each individual sentence of the Wikipedia articles by leveraging the sentence transformers library~\citep{reimers-2019-sentence-bert}. Given an input dialog,  we compute the representation for each turn using an encoder model. The model used to encode the knowledge base \& input dialog is a DistilBert~\citep{sanh2019distilbert} model finetuned on the NLI and STSb tasks.  We take the average of the encoded embeddings and perform kNN search using the FAISS library~\citep{JDH17} for efficiency. In out experiments, we retrieve the top most relevant sentence 
        from the knowledge base.
        \paragraph{(c) Dense Passage Retriever (DPR)}: For the last configuration, we use an open domain DPR retriever model from~\citet{karpukhin2020dense}. In the DPR model, the knowledge base consists of 21 million Wikipedia passages. For our experiments, we use the pretrained encoder model made available from~\citet{karpukhin2020dense} to encode the last turn of the dialog. Using the encoded turn, we retrieve top most relevant passage from the knowledge base by performing a similarity search using FAISS.

To create our annotation dataset we first sample 100 dialogs of length 5 turns from the WoW test sets and obtain responses using each of the configurations described above, resulting in 3600 responses in total (100 dialogs $*$ 4 model sizes $*$ 3 knowledge setups $*$ 3 decoding mechanisms). 
To compute inter-annotator agreement we collect three annotations for each system generated response based on the setup described in Section~\ref{sec:human_annotation}~\footnote{We leverage Mechanical Turk for our annoation https://www.mturk.com/}. 
After the completion of stage 1, we filter out responses whose appropriateness and verifiable scores are below 3 to ensure we have responses that contain some form of knowledge and are relevant to the dialog context. After filtering out inappropriate responses there were 2781 responses remaining. After filtering out non-verifiable responses from this set there were 1869 remaining responses to be annotated in stage 2, that is, 33\% of appropriate responses were non-verifiable. This highlights the significant amount of non-verifiable responses within open-domain dialog. On average 52 responses were annotated during stage 2 for each configuration. We show the exact number responses annotated for each configuration in the Appendix.

For each response we compute the factual consistency score by taking the average across the three annotation scores. For hallucination we do majority voting of the three annotation scores for each response. We use Krippendorff's alpha for inter-annotator agreement (IAA), and found IAA for Factual Consistency is 0.15 and 0.21 for Hallucination indicating slight to fair agreement~\cite{landis1977measurement}. In Section~\ref{sec:experiments} we will create a smaller dataset with higher IAA. 

Table~\ref{human_all_knowledge} shows the annotation results for factual consistency and hallucination under different configurations. Our human annotation results show two insights: we see that none of these state-of-the-art neural generation models generate factually consistent responses all the time and second this rate changes depending on the chosen configuration. As models get larger, there is an increase in factual consistency and a decrease in hallucinated responses.
In a more realistic setting where DPR is being used to retrieve knowledge from a larger knowledge base, the highest average factual consistency score is 0.71, and at the lowest 18.2\% of responses have hallucinated information, indicating room for improvement in factual consistency. Additionally, we see that in the DPR setting, DBS performs well for larger models.~\citet{massarelli2019decoding} showed DBS performed the best for factual verification across all model sizes; however, that was for document-level generation. 

Table~\ref{human_all_knowledge} shows the average appropriateness score from the Stage 1 annotation. We can see that using DPR results in lower appropriateness scores than using the ground truth knowledge as it can return irrelevant knowledge sentences. 
Using NN has higher appropriateness scores than  DPR since the size of knowledge base in NN is smaller (it is from the same set of Wikipedia articles used in WoW), and thus it is more likely to return a relevant knowledge sentence than DPR. 
These results show that the input knowledge quality can affect both the response quality and the factual consistency in model responses.

\section{Factual Consistency Detector}
After demonstrating the factual correctness issues in knowledge-grounded neural response generators,
we develop a model to automatically detect the factual consistency of a generated response with respect to the knowledge fed into a response generation model. Since there are no existing labeled dialog datasets to train such detection models,  we first develop a new dataset, Conv-FEVER, described below, and finetune our model using this dataset.  
We leverage the HuggingFace repo~\cite{wolf-etal-2020-transformers} for all of our experiments.

\subsection{Conv-FEVER Corpus}
\label{sec:conv_fever}
We create the Conv-FEVER dataset from the Wizard of Wikipedia  corpus~\citep{dinan2018wizard} consisting of dialogs between paired crowd sourced workers that played the role of wizard and apprentice in a controlled environment. The conversations between the wizard and apprentice are around particular topics, and the wizard's responses are based on the knowledge sentence chosen from Wikipedia documents about the topic. Therefore we hypothesize that the \textbf{\textit{responses generated by the wizard are consistent with respect to the provided knowledge.}}
To generate inconsistent responses, we leverage a few data augmentation strategies introduced in~\citep{kryscinskiFactCC2019}, including random pairing, negation, entity swapping. Every data point in the Conv-FEVER dataset contains a Conversational Context $(C_n)$, Knowledge $(K_n)$, Wizard's Response or Target Response $(R_n)$, where $n$ represents the size of the dataset. Table~\ref{conv-fever-stats} shows the statistics of the data set. Figure~\ref{fig:data_augmentation} shows examples of these data augmentation techniques.

\begin{table}[!h]
\tabcolsep 3.5pt
\centering
\begin{tabular}{l c c}
 \toprule
\multirow{2}{*}{Dataset} & Num.  & Num.   \\
& Consistent & Inconsistent\\
\hline
WoW & 68957  & - \\
\hdashline[.4pt/1pt]
Random  Pairing & -  & 137914 \\
\hdashline[.4pt/1pt]
Negation & -  & 107845 \\
\hdashline[.4pt/1pt]
Entity & -  & 73178 \\ 

\bottomrule 
\end{tabular}
\caption{Conv-FEVER dataset statistics}
\label{conv-fever-stats}
\end{table}

\paragraph{Random Pairing}: 
we perform two types of random pairing when given a conversational context $(C_i)$, knowledge $(K_i)$, wizard's response $(R_i)$: (1) we replace the response $(R_i)$ associated with conversation ($C_i$) with a response ($R_j$) from a random conversation $(C_j)$ where $j\neq i$; (2) We replace the knowledge $(K_i)$ associated with conversation ($C_i$) with a Knowledge ($K_j$) from a random conversation $(C_j)$ where $j\neq i$.

\paragraph{Negation}: we perform two types of negations: (1) negation on the response whilst keeping the context and knowledge  untouched; (2) negation applied to the knowledge whilst keeping the context and response untouched. In both the cases, we performed negation only if certain verbs\footnote{We apply negation for these tokens: are, is, was, were, have, has, had, do, does, did, can, ca, could, may, might, must, shall, should, will, would.} were present in the response or knowledge. Further as a part of negation, we added or replaced the words with ``not or n't'' as part of negating the sentence.

\paragraph{Entity Swapping}: We performed this data augmentation to ensure that the dataset contains representation of the most common mistakes made by current state of the art generation models. To perform entity swapping, we used SpaCY NER tagger~\cite{spacy} over the conversational context ($C_i$) and knowledge ($K_i$) to create two  indexes of all entities mentioned in context $(E_c)$ and knowledge $(E_k)$ respectively. We then performed two types of entity swapping: (1) we performed entity swapping only on the context if there is a common entity ($Y$) mentioned in the context and the response. In this case, we randomly sampled an entity from the context index $(E_c)$ and replaced the common entity ($Y$); (2) we performed entity swapping only on the knowledge if there is a common entity ($Y$) mentioned in the knowledge and the response. In this case, we randomly sampled an entity from the knowledge index $(E_k)$ and replaced the common entity ($Y$).

\begin{figure}[t]
\centering
\begin{tabular}{p{0.95\linewidth}}
\toprule
{\em Dialog History:}\\
{\bf Speaker 1:} I couldn't imagine living when internet access was rare \\
{\bf Speaker 2:} Oh me either! It seems like such a long time ago. I wonder when Internet was first created? \\ 
\hline
{\em Knowledge}: \\
Use by a wider audience only came in 1995 when restrictions on the use of the Internet to carry commercial traffic were lifted.\\
\hdashline[.4pt/1pt]
{\em Response}:\\ 
It used to be restricted but around 1995, the restrictions were lifted and commercial use of it began\\
\hdashline[.4pt/1pt]
\hdashline[.4pt/1pt]
{\em Negated Response:} \\
It used to be restricted but around 1995, the restrictions ~\textcolor{red}{weren't} lifted and commercial use of it began\\
\hdashline[.4pt/1pt]
{\em Entity Swap Knowledge:} \\
Use by a wider audience only came in~\textcolor{red}{1971} when restrictions on the use of the Internet to carry commercial traffic were lifted.\\
\bottomrule
\end{tabular}
\caption{Data augmentation examples in Conv-FEVER corpus. Red text highlight shows the changes made to the original sentence.}
\label{fig:data_augmentation}
\end{figure}

\begin{table*}[!h]
\centering
\tabcolsep 3.5pt
\begin{tabular}{clcccccccccccc}
\toprule
    \multirow{2}{*}{Retriever} & \multirow{2}{*}{Decoding} && \multicolumn{2}{c}{GPT2-Small} && \multicolumn{2}{c}{GPT2-Medium} && \multicolumn{2}{c}{GPT2-Large} && \multicolumn{2}{c}{GPT2-XL}\\
    \cline{4-5} \cline{7-8} \cline{10-11} \cline{13-14}
    & && F1(C) & F1(IC) && F1(C) & F1(IC) && F1(C) & F1(IC) && F1(C) & F1(IC)\\
    \cline{1-2} \cline{4-5} \cline{7-8} \cline{10-11} \cline{13-14}
    \cline{4-14}
    \multirow{3}{*}{GT} 
    & BS && 86.7 & 0.0 && 87.0 & 0.0 && 86.6  &10.5 && 91.2 & 0.0\\
    & NS && 87.1 & 13.3 && 90.3 & 14.3 && 93.1 & 0.0 &&~\textbf{97.5} & 0.0\\
    & DBS && 87.9 & 31.6 && 89.9 & 0.0 && 93.3 & 50.0 && 93.8 &~\textbf{36.4} \\
    \hdashline[.4pt/1pt]
    \multirow{3}{*}{KNN}
    & BS && 80.5 & 26.1 && 82.7 & 23.5 && 80.0 & 20.0 && 78.9 & 11.8 \\
    & NS && 81.9 & 40.0 && 87.1 & 26.7 &&~\textbf{87.2} & 40.0 && 78.2 & 29.6 \\
    & DBS && 82.4 & 11.8 && 86.8 & 35.3 && 76.5 & 24.0 && 83.1 &~\textbf{43.5} \\
    \hdashline[.4pt/1pt]
    \multirow{3}{*}{DPR}
    & BS && 72.0 & 36.4 && 71.2 & 26.1 && 81.4 & 42.1 && 78.4 & 52.2 \\
    & NS && 66.7 & 26.1 && 78.8 & 30.0 && 80.6 & 22.2 &&~\textbf{84.6} &~\textbf{53.9} \\
    & DBS && 69.4 & 34.8 && 67.9 & 32.0 && 82.9 & 40.0 && 76.0 & 25.0\\ \bottomrule 
\end{tabular}
\caption{\label{tab:factual_large}Results on~\textit{Factual dataset-crowd} for BERT-base Conv-FEVER. F1 (C): F1 (Consistent) ,  F1 (IC): F1 (Inconsistent), BS: Beam-Search, NS: Nucleus Sampling, DBS: Delayed Beam-Search}
\end{table*}


\begin{table}[ht]
        \centering 
        \small
        \tabcolsep 3pt
        \begin{tabular}{ccccccccc} \toprule
         \multirow{2}{*}{Model} && \multicolumn{3}{c}{Consistent} && \multicolumn{3}{c}{Inconsistent}\\
         \cline {3-5} \cline{7-9}
         && P & R & F1 && P & R & F1 \\ \hline
        FactCC && 81.3 & 73.0 & 77.0 && 30.1 & 41.1 & 34.7 \\
        ALBERT + VitC && 93.5 & 50.7 & 65.8 && 33.5 & 87.6 & 48.5 \\
        \hspace{2mm} + FEVER && 92.7 & 53.4 & 67.8 && 34.1 & 85.2 &~\textbf{48.8} \\
        ALBERT + MNLI &&\textbf{95.6} & 43.1 & 59.4 && 31.6 &~\textbf{93.0} & 47.2 \\
        \hspace{2mm} + VitC && 93.2 & 48.5 & 63.8 && 32.5 & 87.6 & 47.5 \\
        \hdashline[.4pt/1pt]
        \begin{tabular}{@{}l@{}}BERT-base \\  CONV-FEVER\end{tabular} && 78.7 &~\textbf{99.8} &~\textbf{88.0} &&\textbf{85.7} &4.60 & 8.80 \\ \bottomrule 
       \end{tabular}
       \caption{Results on~\textit{Factual dataset-expert}. P=Precision, R=Recall.VitC = VitaminC dataset.} \label{tab:cleaner}
\end{table}

\begin{figure}[ht]
\centering
\begin{tabular}{p{0.95\linewidth}}
\toprule 
{\em Dialog History:}\\
{\bf Speaker 1:} Well I mainly make pecan because it is so delicious and usually includes vanilla, and salt to balance the taste \\
{\bf Speaker 2:}  i didnt know vanilla was added into it \\ 
\hline
{\em Knowledge}: Pecan pie Pecan pie is often served with whipped cream, vanilla ice cream, or hard sauce. \\
\hline
{\em Response}: Oh thats interesting! I love Pecan pie, it is often served with whipped cream, vanilla ice cream, or hard sauce. \\
\hdashline[.4pt/1pt]
{\em Ground-truth}: Consistent \\ 
\hdashline[.4pt/1pt]
{\em FactCC model:} Consistent \\
\hdashline[.4pt/1pt]
{\em ALBERT-base-VitaminC model:} Not Enough Info \\
\hdashline[.4pt/1pt]
{\em ALBERT-base-VitaminC-FEVER model:} Not Enough Info \\
\hdashline[.4pt/1pt]
{\em ALBERT-base-MNLI model:} Not Enough Info \\
\hdashline[.4pt/1pt]
{\em ALBERT-base-VitaminC-MNLI model:} Not Enough Info \\
\hdashline[.4pt/1pt]
{\em Bert-base-Conv-FEVER:} Consistent\\
\bottomrule 
\end{tabular}
\caption{Responses can contain spans of verifiable (things pecan pie is served with) and non verifiable (liking pecan pie) information. In this example our detector predicts correctly while the NLI models choose not enough info due to the extra information in the response.}
\label{fig:qualitative_example}
\end{figure}

\subsection{Factual Consistency Detector Models}
\label{sec:detecor model}
\subsubsection{Baseline: ALBERT-base}
For our NLI baseline models we use out-of-the box ALBERT-base~\cite{lan2019albert} models\footnote{https://huggingface.co/tals} that were finetuned on three separate datasets: FEVER~\cite{thorne2018fever}, MNLI dataset~\cite{N18-1101} and VitaminC~\cite{Schuster2019}. 
The MNLI task is an entailment task with three labels: entailment, neutral, and contradiction.
The FEVER and VitaminC tasks are used to verify whether claims are supported by a reference piece of knowledge and have three labels: supports, not enough info, and refutes. 
There is a clear mapping between the labels in such two tasks, i.e., supports corresponds to entailment, not enough info to neutral, and refutes to contradiction.  
Therefore we use these three labels: supports, not enough info, and refutes, from these finetuned baseline models.
The models take in the knowledge along with the response, and predict one of the three output labels.
To match our two classes for factual consistency, we consider the {\it supports} label as factually consistent, and bucket {\it not enough info} and {\it refutes} together as factually inconsistent.
We choose to combine {\it not enough info} examples under factually inconsistent because these are typically cases where the response contains hallucinated information unrelated to the knowledge that does not directly contradict or refute the knowledge.
\subsubsection{Baseline: FactCC}
FactCC~\citep{kryscinskiFactCC2019}, is a BERT-based model used to predict factual consistency for neural summarization models. The model was trained on synthesized negative examples using the same data augmentation methods as described in Section~\ref{sec:conv_fever}. For our experiments we use the open-sourced version of the model.~\footnote{https://github.com/salesforce/factCC}

\subsubsection{BERT-base-Conv-FEVER}
We propose a factual consistency detector using our automatically collected data Conv-FEVER. 
We use the BERT-base model~\citep{devlin2018bert} as the backbone, and initialize it by first training on the FEVER dataset~\citep{thorne2018fever} taken from the set of tasks presented in KILT~\citep{petroni2020kilt}. 
The FEVER task is aimed at determining if a claim can be supported or refuted given a Wikipedia document. Claims that are labeled as supports can be thought of to be consistent and refutes can be thought of be inconsistent. 
To create the initial training corpus, we extracted all data points in the FEVER corpus that contained a pointer to the ground truth Wikipedia documents as knowledge. In total we trained on 48,451 supports and 18,625 refutes\footnote{This number differs from the original since we dropped data points without pointer to Wikipedia documents.}.
Because FEVER is not a conversational dataset, we further finetune the model on Conv-FEVER. The set of hyperparameters used are shown in the Appendix.

\section{Factual Consistency Detector Evaluation}
\label{sec:experiments}

Different knowledge retrieval models can greatly affect response quality as shown in Table~\ref{human_all_knowledge}. Therefore we create a ~\textit{Factual dataset-expert} that only consists of the ground truth knowledge, allowing us to assume the selected knowledge is correct. Additionally due to the lower inter-annotator agreement on \textit{Factual dataset-crowd}, we annotate~\textit{Factual dataset-expert} with three internal annotators.

For this dataset we generate 1000 outputs from the fine-tuned GPT2-small model, using the ground-truth knowledge and nucleus sampling with p=0.9 and follow the same annotation setup described in Section~\ref{sec:human_annotation}.
After filtering responses based on the stage 1 annotations, 584 responses were kept, meaning that 41.6\% were deemed as non-verifiable responses, following a similar pattern as~\textit{Factual dataset-crowd}. After stage 2,  the inter-annotator agreement scores using Krippendorff's alpha are 0.80 and 0.86 on factual consistency and hallucination respectively indicating substantial to almost perfect agreement~\cite{landis1977measurement}. We compute factual consistency for each response by averaging the three scores from the internal annotators. We bucket the generated responses such that a factual consistency score $\geq$ 0.5 is labeled as consistent and $<$ 0.5 is labeled as inconsistent. This results in 502 responses labeled as consistent and 82 responses labeled as inconsistent.

Table~\ref{tab:cleaner} shows the factual detector's performance on~\textit{Factual dataset-expert}.~We see our model trained on Conv-FEVER outperforms just training on a document-level dataset on F1 for the consistent class, while NLI models outperform our model on the inconsistent class. 
This is because the NLI models predict the {\it not enough info} class 58\% of the time, which we bucket into the inconsistent label.

To view our detector's performance under different knowledge retrieval configurations we run our BERT-base Conv-FEVER model on~\textit{Factual dataset-crowd}. We bucket our responses in the same way as~\textit{Factual dataset-expert}, resulting in 1618 consistent responses and 251 inconsistent responses across all configurations.

Table~\ref{tab:factual_large} shows that we achieve similar F1-scores for responses generated by GPT2-small using nucleus sampling with ground truth knowledge, which is the same configuration as~\textit{Factual dataset-expert}. We see the general trend that the overall scores improve for responses generated by larger models but decrease when responses are not generated using the ground truth knowledge. The detector's performance is in line with the appropriateness scores shown in Table~\ref{human_all_knowledge} where the detector performed better when the model generated more relevant responses. This further motivates the need to first filter out non-relevant responses before annotating for factual correctness as it is much harder to determine if a response is factually consistent if it is not first appropriate for that dialog context. Figure~\ref{fig:qualitative_example} shows a qualitative example from our models. We present additional examples in the Appendix.

\section{Conclusions and Future Work}
In this work, we present a case study for factual correctness for knowledge-grounded response generation. We propose a human annotation setup to identify non-verifiable, factually consistent, and hallucinated responses and use this setup to annotate responses from multiple neural generative models. Furthermore, we propose a  detector to identify if a response is factually consistent with its respective input knowledge. We create a new dataset called Conv-FEVER to train a factual consistency detector and through evaluation show the benefit of this dataset. 
Since our factual detector model relies heavily on pretrained models to learn representations for both knowledge and responses, we expect it can generalize well to other knowledge-grounded dialog datasets.

The factual consistency detector can be used to automatically evaluate system responses for factual correctness.  
Therefore our future work involves using factual consistency information as feedback to improve response generation models or utilizing the detector to re-rank responses from a response generator.   Additionally we will look at evaluating factual correctness against the ``world knowledge".

\bibliography{anthology,emnlp2021}

\begin{thebibliography}{41}
\expandafter\ifx\csname natexlab\endcsname\relax\def\natexlab#1{#1}\fi

\bibitem[{Devlin et~al.(2019)Devlin, Chang, Lee, and
  Toutanova}]{devlin2018bert}
Jacob Devlin, Ming-Wei Chang, Kenton Lee, and Kristina Toutanova. 2019.
\newblock \href {https://doi.org/10.18653/v1/N19-1423} {{BERT}: Pre-training of
  deep bidirectional transformers for language understanding}.
\newblock In \emph{Proceedings of the 2019 Conference of the North {A}merican
  Chapter of the Association for Computational Linguistics: Human Language
  Technologies, Volume 1 (Long and Short Papers)}, pages 4171--4186,
  Minneapolis, Minnesota. Association for Computational Linguistics.

\bibitem[{Dinan et~al.(2019)Dinan, Roller, Shuster, Fan, Auli, and
  Weston}]{dinan2018wizard}
Emily Dinan, Stephen Roller, Kurt Shuster, Angela Fan, Michael Auli, and Jason
  Weston. 2019.
\newblock \href {https://openreview.net/forum?id=r1l73iRqKm} {Wizard of
  wikipedia: Knowledge-powered conversational agents}.
\newblock In \emph{7th International Conference on Learning Representations,
  {ICLR} 2019, New Orleans, LA, USA, May 6-9, 2019}. OpenReview.net.

\bibitem[{Du{\v{s}}ek and Kasner(2020)}]{duvsek2020evaluating}
Ond{\v{r}}ej Du{\v{s}}ek and Zden{\v{e}}k Kasner. 2020.
\newblock \href {https://www.aclweb.org/anthology/2020.inlg-1.19} {Evaluating
  semantic accuracy of data-to-text generation with natural language
  inference}.
\newblock In \emph{Proceedings of the 13th International Conference on Natural
  Language Generation}, pages 131--137, Dublin, Ireland. Association for
  Computational Linguistics.

\bibitem[{Dziri et~al.(2021)Dziri, Madotto, Zaiane, and Bose}]{dziri2021neural}
Nouha Dziri, Andrea Madotto, Osmar Zaiane, and Avishek~Joey Bose. 2021.
\newblock Neural path hunter: Reducing hallucination in dialogue systems via
  path grounding.
\newblock \emph{arXiv preprint arXiv:2104.08455}.

\bibitem[{Ghazvininejad et~al.(2018)Ghazvininejad, Brockett, Chang, Dolan, Gao,
  Yih, and Galley}]{ghazvininejad2018knowledge}
Marjan Ghazvininejad, Chris Brockett, Ming-Wei Chang, Bill Dolan, Jianfeng Gao,
  Wen-tau Yih, and Michel Galley. 2018.
\newblock A knowledge-grounded neural conversation model.
\newblock In \emph{Thirty-Second AAAI Conference on Artificial Intelligence}.

\bibitem[{Gopalakrishnan et~al.(2019)Gopalakrishnan, Hedayatnia, Chen,
  Gottardi, Kwatra, Venkatesh, Gabriel, and
  Hakkani-T{\"u}r}]{gopalakrishnan2019topical}
Karthik Gopalakrishnan, Behnam Hedayatnia, Qinlang Chen, Anna Gottardi, Sanjeev
  Kwatra, Anu Venkatesh, Raefer Gabriel, and Dilek Hakkani-T{\"u}r. 2019.
\newblock Topical-chat: Towards knowledge-grounded open-domain conversations.
\newblock \emph{Proc. Interspeech 2019}, pages 1891--1895.

\bibitem[{Holtzman et~al.(2020)Holtzman, Buys, Du, Forbes, and
  Choi}]{holtzman2019curious}
Ari Holtzman, Jan Buys, Li~Du, Maxwell Forbes, and Yejin Choi. 2020.
\newblock \href {https://openreview.net/forum?id=rygGQyrFvH} {The curious case
  of neural text degeneration}.
\newblock In \emph{8th International Conference on Learning Representations,
  {ICLR} 2020, Addis Ababa, Ethiopia, April 26-30, 2020}. OpenReview.net.

\bibitem[{Honnibal et~al.(2020)Honnibal, Montani, Van~Landeghem, and
  Boyd}]{spacy}
Matthew Honnibal, Ines Montani, Sofie Van~Landeghem, and Adriane Boyd. 2020.
\newblock \href {https://doi.org/10.5281/zenodo.1212303} {{spaCy:
  Industrial-strength Natural Language Processing in Python}}.

\bibitem[{Honovich et~al.(2021)Honovich, Choshen, Aharoni, Neeman, Szpektor,
  and Abend}]{honovich2021q}
Or~Honovich, Leshem Choshen, Roee Aharoni, Ella Neeman, Idan Szpektor, and Omri
  Abend. 2021.
\newblock $\uppercase{Q}^{2}$: Evaluating factual consistency in
  knowledge-grounded dialogues via question generation and question answering.
\newblock \emph{arXiv preprint arXiv:2104.08202}.

\bibitem[{Johnson et~al.(2017)Johnson, Douze, and J{\'e}gou}]{JDH17}
Jeff Johnson, Matthijs Douze, and Herv{\'e} J{\'e}gou. 2017.
\newblock Billion-scale similarity search with gpus.
\newblock \emph{arXiv preprint arXiv:1702.08734}.

\bibitem[{Karpukhin et~al.(2020)Karpukhin, Oguz, Min, Lewis, Wu, Edunov, Chen,
  and Yih}]{karpukhin2020dense}
Vladimir Karpukhin, Barlas Oguz, Sewon Min, Patrick Lewis, Ledell Wu, Sergey
  Edunov, Danqi Chen, and Wen-tau Yih. 2020.
\newblock \href {https://doi.org/10.18653/v1/2020.emnlp-main.550} {Dense
  passage retrieval for open-domain question answering}.
\newblock In \emph{Proceedings of the 2020 Conference on Empirical Methods in
  Natural Language Processing (EMNLP)}, pages 6769--6781, Online. Association
  for Computational Linguistics.

\bibitem[{Kryscinski et~al.(2020)Kryscinski, McCann, Xiong, and
  Socher}]{kryscinskiFactCC2019}
Wojciech Kryscinski, Bryan McCann, Caiming Xiong, and Richard Socher. 2020.
\newblock \href {https://doi.org/10.18653/v1/2020.emnlp-main.750} {Evaluating
  the factual consistency of abstractive text summarization}.
\newblock In \emph{Proceedings of the 2020 Conference on Empirical Methods in
  Natural Language Processing (EMNLP)}, pages 9332--9346, Online. Association
  for Computational Linguistics.

\bibitem[{Lan et~al.(2019)Lan, Chen, Goodman, Gimpel, Sharma, and
  Soricut}]{lan2019albert}
Zhenzhong Lan, Mingda Chen, Sebastian Goodman, Kevin Gimpel, Piyush Sharma, and
  Radu Soricut. 2019.
\newblock Albert: A lite bert for self-supervised learning of language
  representations.
\newblock \emph{arXiv preprint arXiv:1909.11942}.

\bibitem[{Landis and Koch(1977)}]{landis1977measurement}
J~Richard Landis and Gary~G Koch. 1977.
\newblock The measurement of observer agreement for categorical data.
\newblock \emph{biometrics}, pages 159--174.

\bibitem[{Li et~al.(2019)Li, Niu, Meng, Feng, Li, and Zhou}]{li2019incremental}
Zekang Li, Cheng Niu, Fandong Meng, Yang Feng, Qian Li, and Jie Zhou. 2019.
\newblock \href {https://doi.org/10.18653/v1/P19-1002} {Incremental transformer
  with deliberation decoder for document grounded conversations}.
\newblock In \emph{Proceedings of the 57th Annual Meeting of the Association
  for Computational Linguistics}, pages 12--21, Florence, Italy. Association
  for Computational Linguistics.

\bibitem[{Logan et~al.(2019)Logan, Liu, Peters, Gardner, and
  Singh}]{logan2019barack}
Robert Logan, Nelson~F. Liu, Matthew~E. Peters, Matt Gardner, and Sameer Singh.
  2019.
\newblock \href {https://doi.org/10.18653/v1/P19-1598} {{B}arack{'}s wife
  hillary: Using knowledge graphs for fact-aware language modeling}.
\newblock In \emph{Proceedings of the 57th Annual Meeting of the Association
  for Computational Linguistics}, pages 5962--5971, Florence, Italy.
  Association for Computational Linguistics.

\bibitem[{Lux et~al.(2020)Lux, Sappelli, and Larson}]{lux2020truth}
Klaus-Michael Lux, Maya Sappelli, and Martha Larson. 2020.
\newblock \href {https://doi.org/10.18653/v1/2020.eval4nlp-1.1} {Truth or
  error? towards systematic analysis of factual errors in abstractive
  summaries}.
\newblock In \emph{Proceedings of the First Workshop on Evaluation and
  Comparison of NLP Systems}, pages 1--10, Online. Association for
  Computational Linguistics.

\bibitem[{Massarelli et~al.(2020)Massarelli, Petroni, Piktus, Ott,
  Rockt{\"a}schel, Plachouras, Silvestri, and Riedel}]{massarelli2019decoding}
Luca Massarelli, Fabio Petroni, Aleksandra Piktus, Myle Ott, Tim
  Rockt{\"a}schel, Vassilis Plachouras, Fabrizio Silvestri, and Sebastian
  Riedel. 2020.
\newblock \href {https://doi.org/10.18653/v1/2020.findings-emnlp.22} {How
  decoding strategies affect the verifiability of generated text}.
\newblock In \emph{Findings of the Association for Computational Linguistics:
  EMNLP 2020}, pages 223--235, Online. Association for Computational
  Linguistics.

\bibitem[{Maynez et~al.(2020)Maynez, Narayan, Bohnet, and
  McDonald}]{maynez-etal-2020-faithfulness}
Joshua Maynez, Shashi Narayan, Bernd Bohnet, and Ryan McDonald. 2020.
\newblock \href {https://doi.org/10.18653/v1/2020.acl-main.173} {On
  faithfulness and factuality in abstractive summarization}.
\newblock In \emph{Proceedings of the 58th Annual Meeting of the Association
  for Computational Linguistics}, pages 1906--1919, Online. Association for
  Computational Linguistics.

\bibitem[{Mielke et~al.(2020)Mielke, Szlam, Boureau, and
  Dinan}]{mielke2020linguistic}
Sabrina~J Mielke, Arthur Szlam, Y-Lan Boureau, and Emily Dinan. 2020.
\newblock Linguistic calibration through metacognition: aligning dialogue agent
  responses with expected correctness.
\newblock \emph{arXiv preprint arXiv:2012.14983}.

\bibitem[{Nan et~al.(2021)Nan, Santos, Zhu, Ng, McKeown, Nallapati, Zhang,
  Wang, Arnold, and Xiang}]{nan2021improving}
Feng Nan, Cicero Nogueira~dos Santos, Henghui Zhu, Patrick Ng, Kathleen
  McKeown, Ramesh Nallapati, Dejiao Zhang, Zhiguo Wang, Andrew~O Arnold, and
  Bing Xiang. 2021.
\newblock Improving factual consistency of abstractive summarization via
  question answering.
\newblock \emph{arXiv preprint arXiv:2105.04623}.

\bibitem[{Petroni et~al.(2020)Petroni, Piktus, Fan, Lewis, Yazdani, De~Cao,
  Thorne, Jernite, Plachouras, Rockt{\"a}schel et~al.}]{petroni2020kilt}
Fabio Petroni, Aleksandra Piktus, Angela Fan, Patrick Lewis, Majid Yazdani,
  Nicola De~Cao, James Thorne, Yacine Jernite, Vassilis Plachouras, Tim
  Rockt{\"a}schel, et~al. 2020.
\newblock Kilt: a benchmark for knowledge intensive language tasks.
\newblock \emph{arXiv preprint arXiv:2009.02252}.

\bibitem[{Petroni et~al.(2019)Petroni, Rockt{\"a}schel, Riedel, Lewis, Bakhtin,
  Wu, and Miller}]{petroni2019language}
Fabio Petroni, Tim Rockt{\"a}schel, Sebastian Riedel, Patrick Lewis, Anton
  Bakhtin, Yuxiang Wu, and Alexander Miller. 2019.
\newblock \href {https://doi.org/10.18653/v1/D19-1250} {Language models as
  knowledge bases?}
\newblock In \emph{Proceedings of the 2019 Conference on Empirical Methods in
  Natural Language Processing and the 9th International Joint Conference on
  Natural Language Processing (EMNLP-IJCNLP)}, pages 2463--2473, Hong Kong,
  China. Association for Computational Linguistics.

\bibitem[{Radford et~al.(2019)Radford, Wu, Child, Luan, Amodei, and
  Sutskever}]{radford2019language}
Alec Radford, Jeffrey Wu, Rewon Child, David Luan, Dario Amodei, and Ilya
  Sutskever. 2019.
\newblock Language models are unsupervised multitask learners.
\newblock \emph{OpenAI blog}, 1(8):9.

\bibitem[{Rashkin et~al.(2021)Rashkin, Reitter, Tomar, and
  Das}]{rashkin2021increasing}
Hannah Rashkin, David Reitter, Gaurav~Singh Tomar, and Dipanjan Das. 2021.
\newblock Increasing faithfulness in knowledge-grounded dialogue with
  controllable features.
\newblock \emph{arXiv preprint arXiv:2107.06963}.

\bibitem[{Reimers and Gurevych(2019)}]{reimers-2019-sentence-bert}
Nils Reimers and Iryna Gurevych. 2019.
\newblock \href {https://doi.org/10.18653/v1/D19-1410} {Sentence-{BERT}:
  Sentence embeddings using {S}iamese {BERT}-networks}.
\newblock In \emph{Proceedings of the 2019 Conference on Empirical Methods in
  Natural Language Processing and the 9th International Joint Conference on
  Natural Language Processing (EMNLP-IJCNLP)}, pages 3982--3992, Hong Kong,
  China. Association for Computational Linguistics.

\bibitem[{Roberts et~al.(2020)Roberts, Raffel, and Shazeer}]{roberts2020much}
Adam Roberts, Colin Raffel, and Noam Shazeer. 2020.
\newblock \href {https://doi.org/10.18653/v1/2020.emnlp-main.437} {How much
  knowledge can you pack into the parameters of a language model?}
\newblock In \emph{Proceedings of the 2020 Conference on Empirical Methods in
  Natural Language Processing (EMNLP)}, pages 5418--5426, Online. Association
  for Computational Linguistics.

\bibitem[{Roller et~al.(2020{\natexlab{a}})Roller, Boureau, Weston, Bordes,
  Dinan, Fan, Gunning, Ju, Li, Poff et~al.}]{roller2020open}
Stephen Roller, Y-Lan Boureau, Jason Weston, Antoine Bordes, Emily Dinan,
  Angela Fan, David Gunning, Da~Ju, Margaret Li, Spencer Poff, et~al.
  2020{\natexlab{a}}.
\newblock Open-domain conversational agents: Current progress, open problems,
  and future directions.
\newblock \emph{arXiv preprint arXiv:2006.12442}.

\bibitem[{Roller et~al.(2020{\natexlab{b}})Roller, Dinan, Goyal, Ju,
  Williamson, Liu, Xu, Ott, Shuster, Smith et~al.}]{roller2020recipes}
Stephen Roller, Emily Dinan, Naman Goyal, Da~Ju, Mary Williamson, Yinhan Liu,
  Jing Xu, Myle Ott, Kurt Shuster, Eric~M Smith, et~al. 2020{\natexlab{b}}.
\newblock Recipes for building an open-domain chatbot.
\newblock \emph{arXiv preprint arXiv:2004.13637}.

\bibitem[{Sanh et~al.(2019)Sanh, Debut, Chaumond, and
  Wolf}]{sanh2019distilbert}
Victor Sanh, Lysandre Debut, Julien Chaumond, and Thomas Wolf. 2019.
\newblock Distilbert, a distilled version of bert: smaller, faster, cheaper and
  lighter.
\newblock \emph{arXiv preprint arXiv:1910.01108}.

\bibitem[{Schuster et~al.(2021{\natexlab{a}})Schuster, Fisch, and
  Barzilay}]{Schuster2019}
Tal Schuster, Adam Fisch, and Regina Barzilay. 2021{\natexlab{a}}.
\newblock \href {https://arxiv.org/abs/2103.08541} {Get your vitamin c! robust
  fact verification with contrastive evidence}.
\newblock In \emph{NAACL 2021}.

\bibitem[{Schuster et~al.(2021{\natexlab{b}})Schuster, Fisch, and
  Barzilay}]{schuster2021get}
Tal Schuster, Adam Fisch, and Regina Barzilay. 2021{\natexlab{b}}.
\newblock Get your vitamin c! robust fact verification with contrastive
  evidence.
\newblock \emph{arXiv preprint arXiv:2103.08541}.

\bibitem[{Shuster et~al.(2021)Shuster, Poff, Chen, Kiela, and
  Weston}]{shuster2021retrieval}
Kurt Shuster, Spencer Poff, Moya Chen, Douwe Kiela, and Jason Weston. 2021.
\newblock Retrieval augmentation reduces hallucination in conversation.
\newblock \emph{arXiv preprint arXiv:2104.07567}.

\bibitem[{Thomson and Reiter(2020)}]{thomson2020gold}
Craig Thomson and Ehud Reiter. 2020.
\newblock \href {https://www.aclweb.org/anthology/2020.inlg-1.22} {A gold
  standard methodology for evaluating accuracy in data-to-text systems}.
\newblock In \emph{Proceedings of the 13th International Conference on Natural
  Language Generation}, pages 158--168, Dublin, Ireland. Association for
  Computational Linguistics.

\bibitem[{Thorne et~al.(2018)Thorne, Vlachos, Christodoulopoulos, and
  Mittal}]{thorne2018fever}
James Thorne, Andreas Vlachos, Christos Christodoulopoulos, and Arpit Mittal.
  2018.
\newblock \href {https://doi.org/10.18653/v1/N18-1074} {{FEVER}: a large-scale
  dataset for fact extraction and {VER}ification}.
\newblock In \emph{Proceedings of the 2018 Conference of the North {A}merican
  Chapter of the Association for Computational Linguistics: Human Language
  Technologies, Volume 1 (Long Papers)}, pages 809--819, New Orleans,
  Louisiana. Association for Computational Linguistics.

\bibitem[{Williams et~al.(2018)Williams, Nangia, and Bowman}]{N18-1101}
Adina Williams, Nikita Nangia, and Samuel Bowman. 2018.
\newblock \href {http://aclweb.org/anthology/N18-1101} {A broad-coverage
  challenge corpus for sentence understanding through inference}.
\newblock In \emph{Proceedings of the 2018 Conference of the North American
  Chapter of the Association for Computational Linguistics: Human Language
  Technologies, Volume 1 (Long Papers)}, pages 1112--1122. Association for
  Computational Linguistics.

\bibitem[{Wolf et~al.(2020)Wolf, Debut, Sanh, Chaumond, Delangue, Moi, Cistac,
  Rault, Louf, Funtowicz, Davison, Shleifer, von Platen, Ma, Jernite, Plu, Xu,
  Scao, Gugger, Drame, Lhoest, and Rush}]{wolf-etal-2020-transformers}
Thomas Wolf, Lysandre Debut, Victor Sanh, Julien Chaumond, Clement Delangue,
  Anthony Moi, Pierric Cistac, Tim Rault, Rémi Louf, Morgan Funtowicz, Joe
  Davison, Sam Shleifer, Patrick von Platen, Clara Ma, Yacine Jernite, Julien
  Plu, Canwen Xu, Teven~Le Scao, Sylvain Gugger, Mariama Drame, Quentin Lhoest,
  and Alexander~M. Rush. 2020.
\newblock \href {https://www.aclweb.org/anthology/2020.emnlp-demos.6}
  {Transformers: State-of-the-art natural language processing}.
\newblock In \emph{Proceedings of the 2020 Conference on Empirical Methods in
  Natural Language Processing: System Demonstrations}, pages 38--45, Online.
  Association for Computational Linguistics.

\bibitem[{Wolf et~al.(2019)Wolf, Sanh, Chaumond, and
  Delangue}]{wolf2019transfertransfo}
Thomas Wolf, Victor Sanh, Julien Chaumond, and Clement Delangue. 2019.
\newblock Transfertransfo: A transfer learning approach for neural network
  based conversational agents.
\newblock \emph{arXiv preprint arXiv:1901.08149}.

\bibitem[{Yavuz et~al.(2019)Yavuz, Rastogi, Chao, and
  Hakkani-Tur}]{yavuz2019deepcopy}
Semih Yavuz, Abhinav Rastogi, Guan-Lin Chao, and Dilek Hakkani-Tur. 2019.
\newblock \href {https://doi.org/10.18653/v1/W19-5917} {{D}eep{C}opy: Grounded
  response generation with hierarchical pointer networks}.
\newblock In \emph{Proceedings of the 20th Annual SIGdial Meeting on Discourse
  and Dialogue}, pages 122--132, Stockholm, Sweden. Association for
  Computational Linguistics.

\bibitem[{Zhou et~al.(2020)Zhou, Neubig, Gu, Diab, Guzman, Zettlemoyer, and
  Ghazvininejad}]{zhou2020detecting}
Chunting Zhou, Graham Neubig, Jiatao Gu, Mona Diab, Paco Guzman, Luke
  Zettlemoyer, and Marjan Ghazvininejad. 2020.
\newblock Detecting hallucinated content in conditional neural sequence
  generation.
\newblock \emph{arXiv preprint arXiv:2011.02593}.

\bibitem[{Zhou et~al.(2018)Zhou, Young, Huang, Zhao, Xu, and
  Zhu}]{zhou2018commonsense}
Hao Zhou, Tom Young, Minlie Huang, Haizhou Zhao, Jingfang Xu, and Xiaoyan Zhu.
  2018.
\newblock Commonsense knowledge aware conversation generation with graph
  attention.
\newblock In \emph{IJCAI}, pages 4623--4629.

\end{thebibliography}
\bibliographystyle{acl_natbib}

\newpage
\appendix

\section{Appendices}
\label{sec:appendix} 

\subsection{Qualitative Examples and Analysis}

Qualitatively, we find that all models is able to correctly identify cases where the response states the fact clearly that is present in the knowledge.

\textit{
\paragraph{Context}: \\
\indent \textbf{Apprentice}: I love going for hikes in nature for exercise. \\
\indent \textbf{Wizard}: It is one of my favorite hobbies as well, Hiking is the preferred term, in Canada and the United States \\
\indent \textbf{Apprentice}: What do they call it elsewhere? \\
\indent \textbf{Wizard}: in the United Kingdom, and the Republic of Ireland, the word "walking" is acceptable to describe "all forms" of walking \\
\indent \textbf{Apprentice}: Well that's interesting, but seems like it could get confusing! I bet humans have been hiking ever since we could walk though! \\
 \hspace{2mm} \\
\noindent \textbf{Knowledge}: Hiking The term bushwalking is endemic to Australia, having been adopted by the Sydney Bush Walkers club in 1927. \\
\hspace{2mm} \\
\noindent \textbf{Response}:
Hiking The term bushwalking is endemic to Australia, having been adopted by the Sydney Bush Walkers club in 1927. \\
\hspace{2mm} \\
\noindent \textbf{Human Annotation}: Consistent \\
\noindent \textbf{FactCC model prediction}: Consistent \\
\noindent \textbf{ALBERT-base-VitaminC model prediction}: Consistent \\
\noindent \textbf{ALBERT-base-VitaminC-FEVER model prediction}: Consistent \\
\noindent \textbf{ALBERT-base-MNLI model prediction}: Consistent \\
\noindent \textbf{ALBERT-base-VitaminC-MNLI model prediction}: Consistent \\
\noindent \textbf{Bert-base-Conv-FEVER prediction}: Consistent \\
}

Responses can contains spans of verifiable (things pecan pie is served with) and non verifiable (liking pecan pie) information. In the example below the our detector predicts correctly while the NLI models choose not enough info due to the extra information in the response.

\textit{
\paragraph{Context}: \\
\indent \textbf{Wizard}: I like to bake pecan pie it is east to make with pecans, eggs, butter, and sugar \\
\indent \textbf{Apprentice}: Im allergic to pecans, unfortunately i cant eat those. i love apple pies though! those are delicious! \\
\indent \textbf{Wizard}: that is terrible apple pie is so tart \\
\indent \textbf{Apprentice}:  What other flavors of pie are there? \\
\indent \textbf{Wizard}: Well I mainly make pecan because it is so delicious and usually includes vanilla, and salt to balance the taste  \\
\indent \textbf{Apprentice}: i didnt know vanilla was added into it \\
\hspace{2mm} \\
\noindent \textbf{Knowledge}: Pecan pie Pecan pie is often served with whipped cream, vanilla ice cream, or hard sauce.\\
\hspace{2mm} \\
\noindent \textbf{Response}:
Oh thats interesting! I love Pecan pie, it is often served with whipped cream, vanilla ice cream, or hard sauce. \\
\hspace{2mm} \\
\noindent \textbf{Human Annotation}: Consistent \\
\noindent \textbf{FactCC model prediction}: Consistent \\
\noindent \textbf{ALBERT-base-VitaminC model prediction}: Not Enough Info \\
\noindent \textbf{ALBERT-base-VitaminC-FEVER model prediction}: Not Enough Info \\
\noindent \textbf{ALBERT-base-MNLI model prediction}: Not Enough Info \\
\noindent \textbf{ALBERT-base-VitaminC-MNLI model prediction}: Not Enough Info \\
\noindent \textbf{Bert-base-Conv-FEVER prediction}: Consistent \\
}

\begin{figure}[t]
 	\centering
 		\includegraphics[width=0.4\textwidth]{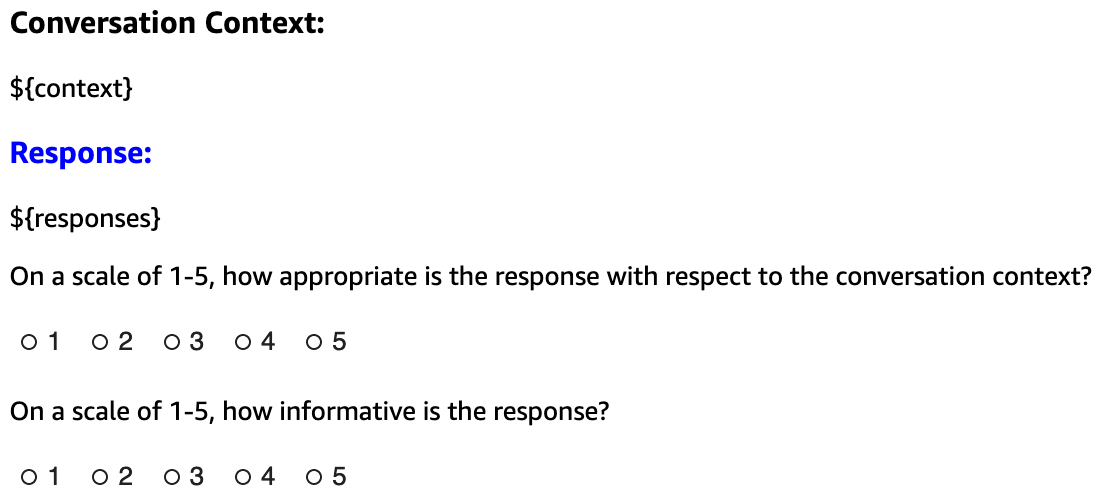}
 	\caption{Interface provided for Stage 1}
	\label{fig:stage_1_interface}
\end{figure}

\begin{figure}[t]
 	\centering
 		\includegraphics[width=0.15\textwidth]{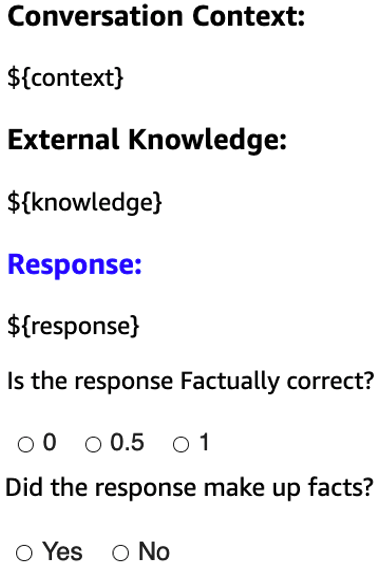}
 	\caption{Interface provided for Stage 2}
	\label{fig:stage_2_interface}
\end{figure}

Example response showcasing hallucination. NLI models predict Not Enough Info which can be thought of as a neutral class. Our detector predicts inconsistent which would be deemed correct as there are no statements in the response that contain any of the knowledge regardless of if it's correct or not. 
\textit{
\paragraph{Context}: \\
\indent \textbf{Apprentice}:  I'm curious to know about Dylan's Candy Bar \\
\indent \textbf{Wizard}: Dylan's candy bar is a chain of boutique candy shop \\
\indent \textbf{Apprentice}: Who founded it? \\
\indent \textbf{Wizard}:  It is owned Dylan Lauren \\ 
\indent \textbf{Apprentice}: What year was it founded? \\
\hspace{2mm} \\
\noindent \textbf{Knowledge}: Dylan's Candy Bar It is owned by Dylan Lauren, daughter of fashion designer Ralph Lauren. \\
\hspace{2mm} \\
\noindent \textbf{Response}:
It was founded in the late 1800s \\
\hspace{2mm} \\
\noindent \textbf{Human Annotation}: Inconsistent \\
\noindent \textbf{FactCC model prediction}: Inconsistent \\
\noindent \textbf{ALBERT-base-VitaminC model prediction}: Not Enough Info \\
\noindent \textbf{ALBERT-base-VitaminC-FEVER model prediction}: Not Enough Info \\
\noindent \textbf{ALBERT-base-VitaminC-MNLI model prediction}: Not Enough Info \\
\noindent \textbf{ALBERT-base-MNLI model prediction}: Not Enough Info \\
\noindent \textbf{Bert-base-Conv-FEVER prediction}: Inconsistent \\
}

We find that our detector does not always capture incorrect dates in responses. The date 2005 does exist in the knowledge but for a different movie. 
\textit{
\paragraph{Context}: \\
\indent \textbf{Apprentice}:  I'm curious to know about Dylan's Candy Bar \\
\indent \textbf{Wizard}: Dylan's candy bar is a chain of boutique candy shop \\
\indent \textbf{Apprentice}: Who founded it? \\
\indent \textbf{Wizard}:  It is owned Dylan Lauren \\ 
\indent \textbf{Apprentice}: What year was it founded? \\
\hspace{2mm} \\
\noindent \textbf{Knowledge}: Jimmy Fallon He left the program for the film industry, starring in films such as "Taxi" (2004) and "Fever Pitch" (2005). \\
\hspace{2mm} \\
\noindent \textbf{Response}:
I have read the book, I love "Taxi", which was released in 2005. \\
\hspace{2mm} \\
\noindent \textbf{Human Annotation}: Inconsistent \\
\noindent \textbf{FactCC model prediction}: Inconsistent \\
\noindent \textbf{ALBERT-base-VitaminC model prediction}: Not Enough Info \\
\noindent \textbf{ALBERT-base-VitaminC-FEVER model prediction}: Not Enough Info \\
\noindent \textbf{ALBERT-base-VitaminC-MNLI model prediction}: Not Enough Info \\
\noindent \textbf{ALBERT-base-MNLI model prediction}: Inconsistent \\
\noindent \textbf{Bert-base-Conv-FEVER prediction}: Consistent \\
}

\begin{table*}[!ht]
\centering 
\tabcolsep 3pt
\begin{tabular}{c c cc c cc c cc c cc}
    \toprule
    Retriever & Decoding & \multicolumn{2}{c}{GPT2-Small} && \multicolumn{2}{c}{GPT2-Medium} && \multicolumn{2}{c}{GPT2-Large} && \multicolumn{2}{c}{GPT2-XL}\\
    \cline{3-4} \cline{6-7} \cline{9-10} \cline{12-13}
    & & F1(C) & F1 (IC) && F1(C) & F1 (IC) && F1(C) & F1 (IC) && F1(C) & F1 (IC)\\
    \midrule 
    \multirow{3}{*}{Ground Truth} 
    & Beam Search & 87.0 & 0.0 && 87.0 & 0.0 && 87.0 & 11.0 && 91.0 & 0.0\\
    & Nucleus Sampling & 87.0 & 13.0 && 90.0 & 14.0 && 93.0 & 0.0 && 97.0 & 0.0\\
    & DBS & 88.0 & 32.0 && 90.0 & 0.0 && 93.0 & 50.0 && 94.0 & 36.0\\ \hdashline[.4pt/1pt]
    \multirow{3}{*}{KNN}
    & Beam Search & 80.0 & 26.0 && 83.0 & 24.0 && 80.0 & 20.0 && 79.0 & 12.0\\
    & Nucleus Sampling & 82.0 & 40.0 && 87.0 & 27.0 && 87.0 & 40.0 && 83.0 & 43.0\\
    & DBS & 82.0 & 12.0 && 87.0 & 35.0 && 77.0 & 24.0 && 78.0 & 30.0\\ \hdashline[.4pt/1pt]
    \multirow{3}{*}{DPR}
    & Beam Search & 72.0 & 36.0 && 71.0 & 26.0 && 81.0 & 42.0 && 78.0 & 52.0\\
    & Nucleus Sampling & 67.0 & 26.0 && 79.0 & 30.0 && 81.0 & 22.0 && 85.0 & 54.0\\
    & DBS & 69.0 & 35.0 && 68.0 & 32.0 && 83.0 & 40.0 && 76.0 & 52.0\\
    \hline
\end{tabular}
\caption{\label{tab:factual_large_bert_base}Results on ~\textit{Factual Dataset-crowd} for Bert-base-Conv-FEVER. F1 (C): F1 Consistent / F1(IC): F1 Inconsistent}
\end{table*}

\begin{table*}[!ht]
\centering 
\tabcolsep 3pt
\begin{tabular}{l l c c c c} \hline
    Retriever & Decoding Strategy & GPT2-small & GPT2-medium & GPT2-large & GPT2-XL\\ \hline
    Ground-Truth & Beam Search & 64 & 65 & 73 & 62\\
    & Nucleus Sampling & 58 & 69 & 70 & 61\\
    & DBS & 63 & 65 & 68 & 62\\ \hdashline[.4pt/1pt]
    \multirow{3}{*}{KNN} & Beam Search & 55 & 46 & 50 & 44\\
    & Nucleus Sampling & 44 & 50 & 57 & 57\\
    & DBS & 51 & 50 & 53 & 50\\ \hdashline[.4pt/1pt]
    DPR & Beam Search & 36 & 41 & 39 & 37\\
    & Nucleus Sampling & 37 & 43 & 45 & 52\\
    & DBS & 36 & 39 & 45 & 33\\ \bottomrule 
\end{tabular}
\caption{\label{stats}
Number of responses annotated for Stage 2 under each knowledge retrieval, model, decoding condition}
\end{table*}

\begin{table*}[!ht]
\centering 
\tabcolsep 3pt
\begin{tabular}{c c cc c cc c cc c cc} \toprule
    Retriever & Decoding & \multicolumn{2}{c}{GPT2-Small} && \multicolumn{2}{c}{GPT2-Medium} && \multicolumn{2}{c}{GPT2-Large} && \multicolumn{2}{c}{GPT2-XL}\\ 
    \cline{3-4} \cline{6-7} \cline{9-10} \cline{12-13}
    & & F1(C) & F1 (IC) && F1(C) & F1 (IC) && F1(C) & F1 (IC) && F1(C) & F1 (IC)\\
    \midrule 
    \multirow{3}{*}{Ground Truth} 
    & Beam Search & 61.0 & 37.0 && 64.0 & 41.0 && 71.0 & 44.0 && 55.0 & 18.0\\
    & Nucleus Sampling & 68.0 & 47.0 && 65.0 & 30.0 && 58.0 & 27.0 && 72.0 & 13.0\\
    & DBS & 63.0 & 38.0 && 64.0 & 35.0 && 67.0 & 35.0 && 59.0 & 19.0\\ \hdashline[.4pt/1pt]
    \multirow{3}{*}{KNN}
    & Beam Search & 61.0 & 55.0 && 31.0 & 34.0 && 55.0 & 53.0 && 42.0 & 30.0\\
    & Nucleus Sampling & 42.0 & 54.0 && 61.0 & 44.0 && 63.0 & 47.0 && 60.0 & 60.0\\
    & DBS & 50.0 & 39.0 && 42.0 & 34.0 && 57.0 & 57.0 && 47.0 & 45.0\\ \hdashline[.4pt/1pt]
    \multirow{3}{*}{DPR}
    & Beam Search & 34.0 & 56.0 && 46.0 & 60.0 && 44.0 & 61.0 && 21.0 & 52.0\\
    & Nucleus Sampling & 38.0 & 67.0 && 49.0 & 53.0 && 44.0 & 44.0 && 39.0 & 52.0\\
    & DBS & 30.0 & 58.0 && 45.0 & 64.0 && 36.0 & 39.0 && 39.0 & 46.0\\
    \hline
\end{tabular}
\caption{\label{tab:factual_large_albert_base_mnli}Results on ~\textit{Factual Dataset-crowd} for ALBERT-base MNLI. F1 (C): F1 Consistent / F1(IC): F1 Inconsistent}
\end{table*}

\begin{table*}[!ht]
\centering 
\tabcolsep 3pt
\begin{tabular}{c c cc c cc c cc c cc} \toprule
    Retriever & Decoding & \multicolumn{2}{c}{GPT2-Small} && \multicolumn{2}{c}{GPT2-Medium} && \multicolumn{2}{c}{GPT2-Large} && \multicolumn{2}{c}{GPT2-XL}\\ 
    \cline{3-4} \cline{6-7} \cline{9-10} \cline{12-13}
    & & F1(C) & F1 (IC) && F1(C) & F1 (IC) && F1(C) & F1 (IC) && F1(C) & F1 (IC)\\ \midrule 
    \multirow{3}{*}{Ground Truth} 
    & Beam Search & 71.0 & 42.0 && 64.0 & 28.0 && 75.0 & 41.0 && 72.0 &  29.0\\
    & Nucleus Sampling & 74.0 & 47.0 && 77.0 & 41.0 && 71.0 & 29.0 && 76.0 & 15.0\\
    & DBS & 69.0 & 44.0 && 67.0 & 29.0 && 75.0 & 40.0 && 70.0 & 28.0\\
    \hdashline[.4pt/1pt]
    \multirow{3}{*}{KNN}
    & Beam Search & 59.0 & 54.0 && 43.0 & 29.0 && 53.0 & 55.0 && 48.0 & 43.0\\
    & Nucleus Sampling & 46.0 & 55.0 && 62.0 & 33.0 && 51.0 & 55.0 && 67.0 & 59.0\\
    & DBS & 61.0 & 47.0 && 59.0 & 36.0 && 57.0 & 47.0 && 54.0 & 50.0\\
    \hdashline[.4pt/1pt]
    \multirow{3}{*}{DPR}
    & Beam Search & 34.0 & 56.0 && 51.0 & 56.0 && 50.0 & 57.0 && 28.0 & 53.0 \\
    & Nucleus Sampling & 50.0 & 43.0  && 41.0 & 62.0 && 50.0 & 43.0 && 52.0 & 56.0 \\
    & DBS & 25.0 & 62.0 && 35.0 & 50.0 && 39.0 & 36.0 && 42.0 & 42.0\\
    \bottomrule \end{tabular}
\caption{\label{tab:factual_large_albert_basic_vitaminc}Results on ~\textit{Factual Dataset-crowd} for ALBERT-base-VitaminC. F1 (C): F1 Consistent / F1(IC): F1 Inconsistent}
\end{table*}

\begin{table*}[!ht]
\centering
\tabcolsep 3pt
\begin{tabular}{c c cc c cc c cc c cc}
    \toprule
    Retriever & Decoding & \multicolumn{2}{c}{GPT2-Small} && \multicolumn{2}{c}{GPT2-Medium} && \multicolumn{2}{c}{GPT2-Large} && \multicolumn{2}{c}{GPT2-XL}\\
    \cline{3-4} \cline{6-7} \cline{9-10} \cline{12-13}
    & & F1(C) & F1 (IC) && F1(C) & F1 (IC) && F1(C) & F1 (IC) && F1(C) & F1 (IC)\\
    \midrule 
    \multirow{3}{*}{Ground Truth} 
    & Beam Search & 72.0 & 43.0 && 58.0 & 32.0 && 65.0 & 35.0 && 65.0 & 21.0\\
    & Nucleus Sampling & 82.0 & 55.0 && 70.0 & 36.0 && 71.0 & 21.0 && 74.0 & 7.0 \\
    & DBS & 70.0 & 48.0 && 56.0 & 25.0 && 68.0 & 36.0 && 68.0 & 22.0\\
    \hdashline[.4pt/1pt]
    \multirow{3}{*}{KNN}
    & Beam Search & 66.0 & 52.0 && 44.0 & 33.0 && 54.0 & 50.0 && 57.0 & 46.0 \\
    & Nucleus Sampling & 75.0 & 68.0 && 73.0 & 47.0 && 63.0 & 47.0 && 71.0 & 61.0 \\
    & DBS &  63.0 & 48.0 && 62.0 & 41.0 && 67.0 & 65.0 && 57.0 & 51.0 \\
    \hdashline[.4pt/1pt]
    \multirow{3}{*}{DPR}
    & Beam Search & 32.0 & 49.0 && 62.0 & 65.0 && 54.0 & 59.0 && 39.0 & 56.0 \\
    & Nucleus Sampling & 29.0 & 57.0 && 57.0 & 43.0 && 46.0 & 38.0 && 56.0 & 52.0 \\
    & DBS & 37.0 & 62.0 && 54.0 & 59.0 && 43.0 & 37.0 && 47.0 & 44.0\\
    \bottomrule 
\end{tabular}
\caption{\label{tab:factual_large_albert_base_vitaminc_fever}Results on ~\textit{Factual Dataset-crowd} for ALBERT-base-VitaminC-FEVER. F1 (C): F1 Consistent / F1(IC): F1 Inconsistent}
\end{table*}

\begin{table*}[!ht]
\centering
\tabcolsep 3pt
\begin{tabular}{c c cc c cc c cc c cc}
    \toprule
    Retriever & Decoding & \multicolumn{2}{c}{GPT2-Small} && \multicolumn{2}{c}{GPT2-Medium} && \multicolumn{2}{c}{GPT2-Large} && \multicolumn{2}{c}{GPT2-XL}\\
    \cline{3-4} \cline{6-7} \cline{9-10} \cline{12-13}
    & & F1(C) & F1 (IC) && F1(C) & F1 (IC) && F1(C) & F1 (IC) && F1(C) & F1 (IC)\\
    \midrule 
    \multirow{3}{*}{Ground Truth} 
    & Beam Search & 60.0 & 33.0 && 62.0 & 37.0 && 69.0 & 43.0 && 54.0 & 22.0 \\
    & Nucleus Sampling & 75.0 & 51.0 && 69.0 & 36.0  && 56.0 & 24.0 && 70.0 & 13.0 \\
    & DBS & 67.0 & 46.0 && 58.0 & 26.0 && 67.0 & 38.0 && 55.0 & 18.0 \\
    \hdashline[.4pt/1pt]
    \multirow{3}{*}{KNN}
    & Beam Search & 67.0 & 55.0 && 43.0 & 29.0 && 48.0 & 48.0 && 47.0 & 39.0 \\
    & Nucleus Sampling & 57.0 & 59.0 && 63.0 & 45.0 && 67.0 & 49.0 && 68.0 & 62.0 \\
    & DBS & 51.0 & 43.0 && 54.0 & 34.0 && 58.0 & 59.0 && 51.0 & 49.0 \\
    \hdashline[.4pt/1pt]
    \multirow{3}{*}{DPR}
    & Beam Search & 39.0 & 54.0 && 54.0 & 62.0 && 38.0 & 57.0 && 44.0 & 57.0 \\
    & Nucleus Sampling & 37.0 & 64.0 && 49.0 & 53.0 && 40.0 & 40.0 && 43.0 & 53.0 \\
    & DBS & 30.0 & 58.0 && 55.0 & 67.0 && 43.0 & 41.0 && 42.0 & 42.0 \\
    \bottomrule 
\end{tabular}
\caption{\label{tab:factual_large_albert_base_vitaminc_mnli}Results on ~\textit{Factual Dataset-crowd} for ALBERT-base-VitaminC-MNLI. F1 (C): F1 Consistent / F1(IC): F1 Inconsistent)}
\end{table*}

\begin{table*}[!ht]
\centering 
\tabcolsep 3pt
\begin{tabular}{c c cc c cc c cc c cc} \toprule
    Retriever & Decoding & \multicolumn{2}{c}{GPT2-Small} && \multicolumn{2}{c}{GPT2-Medium} && \multicolumn{2}{c}{GPT2-Large} && \multicolumn{2}{c}{GPT2-XL}\\
    \cline{3-4} \cline{6-7} \cline{9-10} \cline{12-13}
    & & F1(C) & F1 (IC) && F1(C) & F1 (IC) && F1(C) & F1 (IC) && F1(C) & F1 (IC)\\
    \midrule  \multirow{3}{*}{Ground Truth} 
    & Beam Search & 67.0 & 26.0 && 73.0 & 19.0 && 85.0 & 48.0 && 70.0 & 18.0 \\
    & Nucleus Sampling & 75.0 & 21.0 && 85.0 & 26.0 && 78.0 & 25.0 && 83.0 & 0.0 \\
    & DBS & 69.0 & 31.0 && 75.0 & 14.0 && 83.0 & 40.0 && 66.0 & 17.0 \\
    \hdashline[.4pt/1pt] \multirow{3}{*}{KNN}
    & Beam Search & 68.0 & 48.0 && 62.0 & 14.0 && 67.0 & 26.0 && 73.0 & 18.0 \\
    & Nucleus Sampling & 62.0 & 21.0 && 79.0 & 33.0 && 74.0 & 27.0  && 72.0 & 30.0 \\
    & DBS & 67.0 & 30.0 && 69.0 & 21.0 && 71.0 & 44.0 && 60.0 & 18.0\\
    \hdashline[.4pt/1pt] \multirow{3}{*}{DPR}
    & Beam Search & 70.0 & 55.0 && 71.0 & 52.0 && 56.0 & 46.0 && 45.0 & 35.0 \\
    & Nucleus Sampling & 71.0 & 55.0 && 72.0 & 32.0 && 69.0 & 34.0 && 76.0 & 48.0 \\
    & DBS & 50.0 & 50.0 && 64.0 & 45.0 && 48.0 & 22.0 && 60.0 & 26.0 \\
    \bottomrule \end{tabular}
\caption{\label{tab:factual_large_factcc}Results on ~\textit{Factual Dataset-crowd} for FactCC. F1 (C): F1 Consistent / F1(IC): F1 Inconsistent}
\end{table*}

\subsection{Human Evaluation Interface}

We include images of the interface provided to crowdworkers for our human annotation in figures \ref{fig:stage_1_interface} and \ref{fig:stage_2_interface}.

\subsection{Response Generation Hyperparameters}
We finetune the GPT2 model on the Wizard of Wikipedia dataset using dialog context and knowledge as input. We train the both the language modeling and multiple choice head of GPT2 in a TransferTransfo fashion where the loss from each head is weighted equally. We truncate the dialog context to use the past 128 tokens and we truncate the knowledge to use the first 64 tokens. We use ADAM as an optimizer with a learning rate of 6.25e-05 and a piecewise linear scheduler. We use a batch size of 2 and train for 10 epochs using perplexity for early stopping with a patience of 1.

\subsection{BERT-base Conv-FEVER model Hyperparameters}
We finetune the BERT-base-uncased model on the Conv-FEVER dataset using knowledge and response as input. We use ADAM as an optimizer with a learning rate of 2e-05. We use a batch size of 16 examples and train for 4 epochs.

\subsection{Albert-base model Hyperparameters}
All models were used out-of-the box and taken from the HuggingFace repo~\footnote{https://huggingface.co/tals}. These models were trained for 50000 max steps using a batch size of 32 examples and a learning rate of 2e-5. 

\subsection{FactCC model Hyperparameters}
This model was taken out-of-the box~\footnote{https://github.com/salesforce/factCC}. The model was trained for 10 epochs using batch size of 12 examples and learning rate of 2e-5.

\subsection{Factual Consistency Detector Evaluation}

In table \ref{tab:factual_large_bert_base}, 
\ref{tab:factual_large_albert_base_mnli},
\ref{tab:factual_large_albert_basic_vitaminc}, 
\ref{tab:factual_large_albert_base_vitaminc_fever},
\ref{tab:factual_large_albert_base_vitaminc_mnli} and 
\ref{tab:factual_large_factcc} we present results on ~\textit{Factual Dataset-crowd} trained on various Bert or Albert based models trained on FEVER, Conv-Fever, MNLI, VitC or a combination of them and FactCC.

\end{document}